\documentclass[conference]{IEEEtran}
\IEEEoverridecommandlockouts
\usepackage{cite}
\usepackage{amsmath,amssymb,amsfonts}
\usepackage{algorithmic}
\usepackage{graphicx}
\usepackage{multirow}
\usepackage{hhline}
\usepackage{dirtytalk}
\usepackage{textcomp}
\usepackage{xcolor}
\usepackage{csquotes}
\usepackage{comment}
\usepackage{soul}
\usepackage{ulem}
\usepackage{enumitem}
\usepackage{todonotes}

\def\BibTeX{{\rm B\kern-.05em{\sc i\kern-.025em b}\kern-.08em
  T\kern-.1667em\lower.7ex\hbox{E}\kern-.125emX}}
\begin{document}

\title{minOffense: Inter-Agreement Hate Terms for \\ Stable Rules, Concepts, Transitivities, and Lattices}
\vspace{-25pt}

\author{\hspace{-3cm}\IEEEauthorblockN{Animesh Chaturvedi}
\IEEEauthorblockA{\hspace{-3cm}\textit{Indian Institute of Information Technology Dharwad}\\
\hspace{-3cm}Dharwad, Karnataka, (India)\\ \hspace{-3cm}animesh.chaturvedi88@gmail.com}

\and

\IEEEauthorblockN{Rajesh Sharma}
\IEEEauthorblockA{\textit{University of Tartu}\\
Tartu, (Estonia)\\
rajesh.sharma@ut.ee}
}

\maketitle

\begin{abstract}
Hate speech classification has become an important problem due to the spread of hate speech on social media platforms. For a given set of Hate Terms lists (HTs-lists) and Hate Speech data (HS-data), it is challenging to understand which hate term contributes the most for hate speech classification. This paper contributes two approaches to quantitatively measure and qualitatively visualise the relationship between co-occurring Hate Terms (HTs). Firstly, we propose an approach for the classification of hate-speech by producing a Severe Hate Terms list (\textit{Severe HTs-list}) from existing HTs-lists. To achieve our goal, we proposed three metrics (\textit{Hatefulness}, \textit{Relativeness}, and \textit{Offensiveness}) to measure the severity of HTs. These metrics assist to create an \textit{Inter-agreement HTs-list}, which explains the contribution of an individual hate term toward hate speech classification. Then, we used the Offensiveness metric values of HTs above a proposed threshold \textit{minimum Offense (minOffense)} to generate a new \textit{Severe HTs-list}. To evaluate our approach, we used three hate speech datasets and six hate terms lists. Our approach shown an improvement from 0.845 to 0.923 (best) as compared to the baseline. Secondly, we also proposed \textit{Stable Hate Rule} (SHR) mining to provide ordered co-occurrence of various HTs  with \textit{minimum Stability (minStab)}. The SHR mining detects frequently co-occurring HTs to form \textit{Stable Hate Rules} and \textit{Concepts}. These rules and concepts are used to visualise the graphs of \textit{Transitivities} and \textit{Lattices} formed by HTs.

\textbf{
WARNING: This paper contains offensive language that commonly appears in hate speeches. Hate speech text present in this paper does not represent the views of authors.
}

\end{abstract}

\begin{IEEEkeywords}
Harmful Content Online, Hate Speech, Natural Language Processing, Computational Linguistics, Data Analytics.
\end{IEEEkeywords}

\vspace{-5pt}
\section{Introduction}

The spread of hate speech, especially on online social media platforms has forced Governments and policy makers to enact laws to curb this menace. Researchers have presented various solutions for predicting hate speech. These solutions often include the usage of predefined set of Hate Terms (HTs) (e.g., f*ggot, b*tch, f*ck, etc.) for identifying hate speech. However, these solutions miss the fact that for a given Hate Speech data (HS-data) a list of HTs might not be effective in detecting Hate-Speech effectively. This is due to the fact that a Hate Terms list (HTs-list) might not be able to capture the severity and context of a HT, which is always important in detecting Hate-Speech. For multiple HS-data and multiple HTs-lists, we aim to investigate: i) quantitative and ii) qualitative analysis.

Firstly, we perform quantitative analysis \cite{vidgen2019challenges} using three proposed metrics: \textit{Hatefulness}, \textit{Relativeness}, and \textit{Offensiveness}. (See Section \ref{qtAna1}). These metrics are inspired by the concepts of Shapley value \cite{shapley1953notes}\cite{roth1988shapley}, in particular on the idea of ``the contribution by individual players in a game". In our case, these metrics calculate ``the contribution of an individual HT towards hate speech", which is then used for generating Severe HTs-list for a specific HS-data.

Secondly, we perform qualitative analysis inspired from Association Rule Mining, which provides an exhaustive list of association rules as an output. To extract only important rules, the "\textit{interestingness}" measure \cite{geng2006interestingness} uses multiple thresholds to retrieve interesting and significant rules out of all the exhaustive rules. The interestingness separates interesting rules from the less or non interesting rules. The occurrence of any two HTs, for example, $A$ and $B$ together (where $A$ $\rightarrow$ $B$) can have three interestingness thresholds:

\noindent \textit{1) minimum Support (\textbf{minSup})} is a threshold for minimum number of occurrences of HTs $A$ and $B$ occurring together,

\noindent \textit{2) minimum Confidence (\textbf{minConf})} is a threshold for minimum number of occurrences of $A$ $\cup$ $B$ divided by number of occurrences of HT $A$ i.e., \begin{math} \textbf{N($A$ } \cup \textbf{ $B$)} \div \textbf{N($A$)} \end{math}.

\noindent \textit{3) minimum Stability (\textbf{minStab})} is a threshold for minimum number of states in which rule exceeds minSup \& minConf \cite{chaturvedi2019minstab} -\cite{Chaturvedi2022SysNetAnalytics}. In  our case, the stability is the number of HS-data in which a \textit{hate rule} occurs with sufficient minSup and minConf. Hate rule occurring more than a minStab number are said to be \textit{Stable Hate Rule}. 

To help in making efficient HS decisions, this work investigates the following four research questions (RQs).

\textbf{RQ1:} How to perform \textit{Inter-agreement analysis}, which provide information about common HTs between a HS-data and multiple HTs-lists? 

\textbf{Approach:} We generate an \textit{Inter-agreement HTs-list} that contains HTs and two kinds of information about these HTs. First, it contains Offensiveness metric value of each HT in the HS-data and second, the HTs-lists which contains those HTs. The Inter-agreement HTs-list means agreement between the HS-data and the multiple HTs-lists (see Section II). 

\textbf{RQ2} How to use an Inter-agreement HTs-list to generate a \textit{Severe HTs-list} for efficient Hate Speech classification? 

\textbf{Approach:} From the Inter-Agreement HTs-list, we generate the Severe HTs-list having HTs with Offensiveness values above a proposed threshold \textit{minimum Offense (minOffense)}. Severe HTs-list helps in HS classification (see Section II).

\textbf{RQ3} How much better classification is achieved using the Severe HTs-list compared to any of the given HTs-lists? 

\textbf{Approach:} We made a confusion-matrix to demonstrate the quality of classification for a HS-data over classes (\textit{Hate}, \textit{Relative-hate}, \textit{No-hate}). We found Severe HTs-list provides better results for confusion-matrix (TP, TN, FP, FN, precision, recall, f-measure, and accuracy). This provides empirical justification, to measure the inter-agreement between the Severe HTs-list with a HS-data as compared to the inter-agreement between the given set of HTs-lists with the HS-data (see Section II).

\textbf{RQ4a:} How to generate Stable Hate Rules (SHRs) that represent frequently co-occurring HTs among multiple HS-data? \textbf{Q4b} How to make hate concepts and visualise the relationship between co-occurring HTs from SHRs?

\textbf{Approach:} The Inter-agreement HTs-list is used to make an intermediate representational database. Then, we used Stable Hate Rules mining over the database to discover co-occurring HTs. This helps to discover and analyse the co-occurring concepts of HTs (see Section III).


By exploring the solutions to the above four research questions, we put forward a mechanism for effective analysis: severity of each HT and co-occurrences of HTs. Note, the notation N denotes total Number of terms, lists, or datasets etc. To the best of our knowledge, this work is the first to introduce the metrics that measures the severity of Hate Terms.
\section{Quantitative analysis: Inter-Agreement and Severe Hate Terms lists}\label{qtAna1}

This section provides a quantitative analysis approach to generate an Inter-agreement HTs-list, which is further used to extract the Severe HTs-list for efficient hate speech classification. It is to be noted that HS-data may include non-hate data, either annotated wrongly as hate or misclassified as hate due to an inefficient classifier. In addition, we define the following three classes of hate speech, which captures the intensity of hate being present in the hate speech

\noindent \textbf{-} \textbf{Hate:} class indicates the lines definitely contain HTs.

\noindent \textbf{-} \textbf{Relative-hate:} class indicates the lines contain mild HTs.

\noindent \textbf{-} \textbf{No-hate:} class indicates the lines do not contain HTs.



We first describe hate speech analysis based on a HS-data with a single HTs-list (Section \ref{sec:SingleHTList}), then we explain a HS-data analysis using multiple HTs-lists (Section \ref{sec:MultipleHTList}).

\subsection{Single Hate Terms List Analysis}\label{sec:SingleHTList}
To do HS-data analysis, we retrieved \textbf{five intra-agreement artifacts} using a single HTs-list. We analyse HTs (in HTs-list) appearing in a line (e.g. in tweet or post) based on three classes (\textit{Hate}, \textit{Relative-hate}, or \textit{No-hate)}.

\textbf{1. Creation of hate terms frequencies:} As a first step, we parse the whole HS-Data text into separate lines based on the number of HTs it contains. It is denoted by $N$(X), where X $\in$ [1,$Z$] (where $Z$ represents the maximum number of hate terms in a line) and for a certain X value $N$(X) points to all the lines which contain $X$ number of HTs. Some X may not exist in N(X) i.e, X might be a discontinuous series of integers.
    
\textbf{2. AllHateTermsFrequencies and TopTermsFrequency:} For each class of HS-data, these two provides information about the number of each hate term's occurrences (i.e. HTs' frequencies) and the HTs with top-X frequencies (where $X$ is an input parameter), respectively across the whole HS-Data. 
     
    
\textbf{3. AllHTsPercentLine:} For each HT in each class, the percentage of lines containing the HT across all the lines belonging to that class. 
     
\textbf{4. OuterJoinHTsFrequencies and OuterJoinHTsPercentLines:} These two give information about each HT (with frequency and its percentage) occurrences in a HS-data class. \textit{AllHateTermsFrequencies} of all HTs in a class is used to produce outer join of frequencies of HTs in all classes. \textit{AllHTsPercentLine} of each class is used to produce outer join of percentage HS-lines where HTs occur in all classes.
     
\textbf{5. Intra-Agreement between a single HTs-list and a HS-data:} This provides the contribution of HTs in a single HTs-list towards a HS-data. The contribution is measured using two proposed metrics: Hatefulness and Relativeness.

For each HTs over three classes (Hate, Relative-hate, and No-hate), we calculate Hatefulness and Relativeness. These metrics have two cases for i) Hate class, and ii) Hate and Relative-hate classes.

\textbf {Definition: Hatefulness} would be 1 if the Hate Term (HT) occurs in the hate speech class, otherwise it would be 0. Mathematically, 
\vspace{-5pt}
\begin{equation*}
\text{Hatefulness = }
\{ \text{1 or 0 } | \text{ HT}\in\text{Hate class or not, respectively} \}
\end{equation*}

\textbf{Definition: Relativeness} is the proportion of hate class to other classes (Relative-hate or No-hate). Mathematically,
Relativeness (Hate) = 
\vspace{-2pt}
\begin{equation*}
\frac{\text{FreqHT in Hate Class}}{\text{FreqHT in Relative-hate class and FreqHT in No-hate class}}
\end{equation*}
Relativeness (Hate + Relative-hate) = 
\vspace{-2pt}
\begin{equation*}
\frac{\text{FreqHT in Hate Class + FreqHT in Relative-hate class}}{\text{FreqHT in No-hate class}}
\end{equation*}
where FreqHT is the number of occurrences of the HT. In relativeness, when both numerator and denominator provide zero, it reflects HT neither belongs to Hate nor it belongs to No-hate classes in that HS-data. Relative-hate class leads to the ambiguous situation, where hate is contextually hateful.

\subsection{Multiple Hate Terms Lists analysis}\label{sec:MultipleHTList}

Mostly, the current state-of-the-art is limited to the hate speech analysis using single HTs-list. Next, we extend single HTs-list based analysis to multi HTs-lists analysis. 

\textbf{Inter-agreement Hate Terms Analysis} defined as the analysis of a HS-data, which is performed using a given set of multiple HTs-lists = \{HTs-list1, HTs-list2,... HTs-listN\}. Mathematically, we describe Inter-Agreement HTs Analysis as a matrix IA of size \(N \times M\), where N represents the number of HTs-lists and M number of classes in a HS-data. The \(\text{IA}_\text{ij}\) represents the information about HTs of a given HTs-lists, which are present in a class of HS-data. Here, i represents HTs-list and j represents HS-data class. This answers \textbf{RQ1}.

In addition to the five intra-agreement artifacts, to perform the Inter-agreement HTs analysis we also created \textbf{three inter-agreement artifacts} that are defined as follows.

\textbf{6. Inter-Agreement HTs:} For a given HS-data, this gives information about three metrics (Hatefulness, Relativeness, and Offensiveness) of all HTs present across all HTs-lists. This artifact also provides information about membership of HT ``which HTs-list contains which HT''. This list is used to generate \textit{Severe HTs-list} with significantly severe HTs.

\begin{figure}[htbp]
\vspace{-5pt}
\includegraphics[width=1.0 \linewidth]{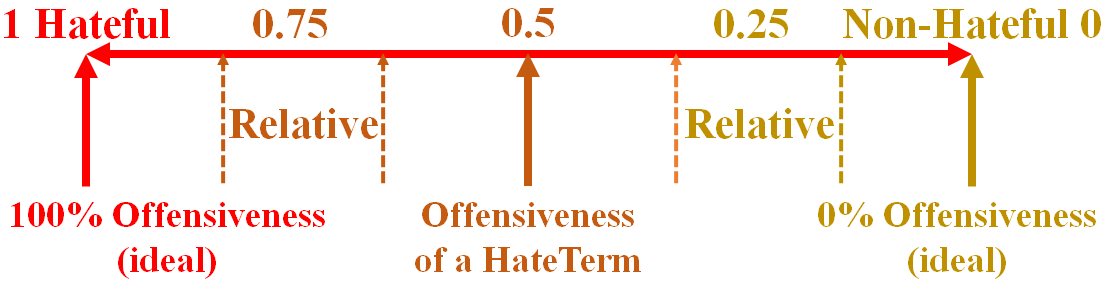}
\vspace{-10pt}
\caption{Varying value of the Offensiveness of a HT for a given HS-data divided into classes (Hate, Relative-hate, and No-hate).}
\label{Fig1}
\end{figure}

We defined the contribution of a HT towards the three labels (Hate, Relative-hate, No-hate) to generate the \textit{Intra-agreement HTs Matrix}. For a matrix, let each HT have some contribution in HS-data due to Hatefulness and Relativeness. The Fig. \ref{Fig1} demonstrates the varying value of this Offensiveness for a HT. For a given HS-data, the HT will be most Hateful when its Offensiveness equals to 1 and the HT will be least hateful when its Offensiveness value is equal to 0.

\textbf{Definition: Offensiveness} is the Harmonic mean of the \textit{Hatefulness} and \textit{Relativeness}. Another alternative means could be the Geometric mean. Mathematically,  
\begin{equation*}
\text{Offensiveness = }
\frac{\text{2 } \times \text{ Hatefulness } \times \text{Relativeness}}{\text{(Hatefulness + Relativeness)}} 
\end{equation*}

For each HT, the calculated Offensiveness signifies the severity of a hate term. The severity of the hate term is the interestingness with high Offensiveness value. For hate-severity of a HT, we formulate the Offensiveness to classify terms either as hate or non-hate. This means, we considered the percentage contribution (i.e., overall input towards cost) of a hate term occurrences to a hate class. The Hatefulness, Relativeness, and Offensiveness provide severity of HTs using HS-data labels. The three metrics are calculated for all HTs occurring in a given set of HTs-lists.

\textbf{Answers to RQ2}. The steps to generate \textit{Severe HTs-list} using Inter-agreement HTs-list, which assists in better inter-agreement analysis. The Offensiveness provides help to separate out the highly severe HTs and the less severe HTs. Based on the high values of Offensiveness, we generated the Severe HTs-lists. The Severe HTs-list has been generated with HTs having Offensiveness value greater than a user-defined interestingness threshold \textbf{minimum Offense (minOffense)}. This step can be performed iteratively and heuristically to generate the Severe HTs-lists with various minOffense values. The expectation of the Severe HTs-list is to produce better hate speech classification as compared to the given set of HTs-lists. We present our results in a confusion-matrix as the Inter-Agreement between the HS-data and the given set of HTs-lists.

\begin{figure}[htbp]
\includegraphics[width=1.0 \linewidth]{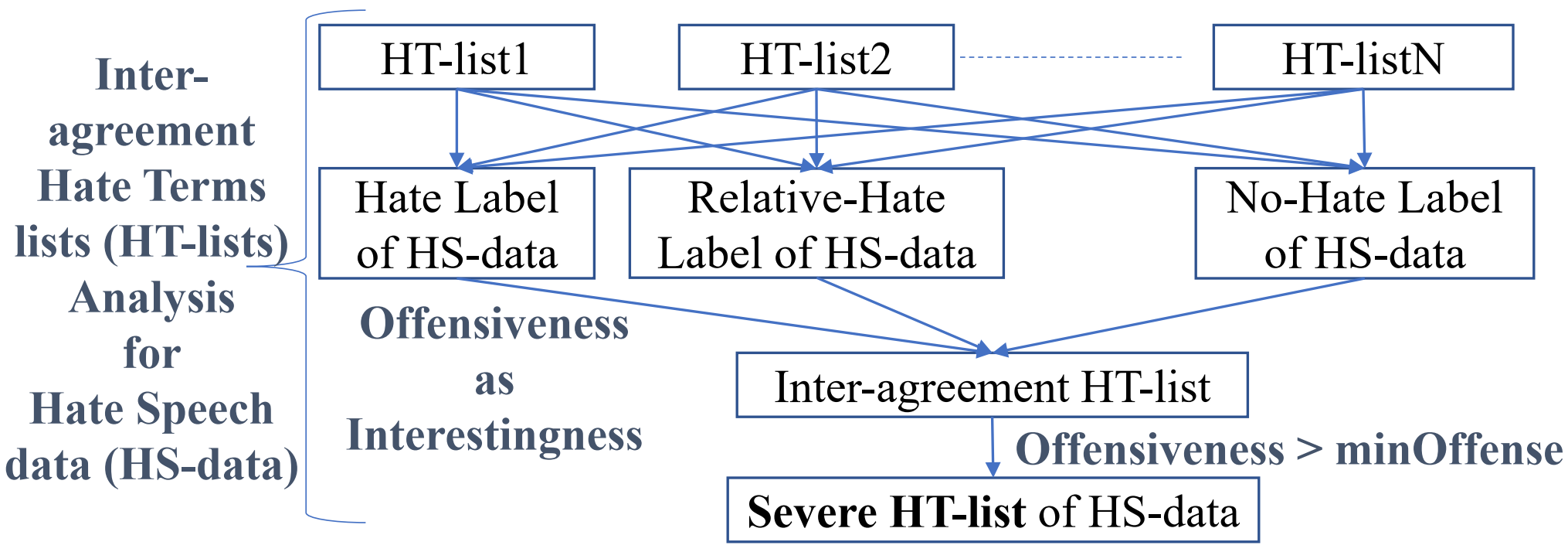}
\vspace{-10pt}
\caption{Flow of Inter-agreement analysis for generation of Severe HTs-list.}
\label{Fig2}
\vspace{-7pt}
\end{figure}

\textbf{7. Inter-agreement Confusion-matrix:} This provides information about confusion-matrix with True Positive (TP), True Negative (TN), False Positive (FP), and False Negative (FN) for the calculation of accuracy, precision, recall, and f-measure of each HT in all HTs-list for a HS-data. The calculation is done for the three cases given in the Table \ref{Tab1}. 

\begin{table*}
\centering
\caption{Confusion-matrix of three Case-studies.}\label{Tab1}
\begin{tabular}{|p{2cm}|p{3.7cm}|p{3.4cm}|p{3.6cm}|p{3.3cm}|}
\hline
\textbf{CaseStudy class} & \textbf{TP} = percentage of HS-lines & \textbf{TN} = percentage of HS-lines & \textbf{FP} = percentage of HS-lines & \textbf{FN} = percentage of HS-lines \\
\hline
\textbf{Hate} & with HTs occurring in Hate class & without HTs occurring in NonOffensive class & with HTs in NonOffensive class & without HTs in Hate class \\
\hline
\textbf{Relative-hate} & with HTs occurring in Offensive class & without HTs occurring in NonOffensive class & with HTs in NonOffensive class & without HTs in Offensive class \\
\hline
\textbf{Hate + Relative-hate} & with HTs occurring in Hate+Offensive class & without HTs occurring in NonOffensive class & with HTs in NonOffensive class & without HTs in Hate+Offensive class \\ 
\hline
\end{tabular}
\vspace{-7pt}
\end{table*}

As per binary-relevance \cite{zhang2018binary}, we combine two HS-data classes (Hate, Relative-hate, No-hate) only if they are correlated. This is useful, when it is required to model hate-speech analysis as binary classification. Thus, we transformed three classes to binary classes, where two similar classes (e.g. Hate and Relative-hate) represent one class versus (Vs) No-hate. This means, we can combine Hate with Relative-hate assuming both are binary relevant. Thus, we exploited the binary relevance (i.e., correlation) to do binary classification denoted as Hate + Relative-hate Vs No-hate.

For the three classes, there are a total of $8$ ($2^3$) possible cases. Considering binary-relevance, the Relative-hate class can fall into Hate or No-hate. This reduce the number of cases to 6:
Hate Vs No-hate; Hate Vs Relative-hate; Hate Vs Relative-hate + No-hate; Hate + Relative-hate Vs No-hate; Relative-hate Vs No-hate; No-hate Vs Hate + Relative-hate.

In a HS-data (corpus), generally it is observed that occurrence of hateful speech is less as compared to non-hateful speech. This leads to well-known imbalanced classes (Hate Vs. No-hate) problems based on frequency of HS-lines that contain HT or not-contains HTs. For example, Davidson et al. \cite{davidson2017automated} defined the three classes: hate, offensive, and non-offensive, which has imbalanced class, as Hate class has 1431 tweets, 19191 tweets for Offensive, and 4164 for Non-offensive class. In this condition, the \textit{frequency of classified tweets} represents an imbalanced confusion-matrix (TP, TN, FP, FN) in the three classes. Hence, we need to use the \textit{percentage (or ratio) of classified tweets} with respect to ``size of a particular dataset''. The \textit{percentage of classified HS-lines} are used to represent TP, TN, FP, FN as given in Table \ref{Tab1} for proper scaling. For example, 1400 hates are correctly classified as TP in Hate class, which is smaller than 5000 incorrectly classified as FP in Offensive class. However, 1400/1431 = 0.97 hate in Hate class for TP and 5000/19191 = 0.26 hate in Offensive class for FP, here ratios are properly scaled according to the class size. The percentage of HS-lines in a given class is a better measure. In the Table \ref{Tab1}, we used the percentage of HS-lines in a class to evaluate metrics, which gave a fair result by avoiding imbalance. The results of the Severe HTs-list are compared with the given set of HTs-lists by common binary metrics: accuracy, precision, recall, and F-measure (given in Table IX).

\textbf{Answers to RQ3:} For a HS-data, we discovered two facts. Fact 1: for best recall, the FN should be zero. This happens when all HTs (in a HTs-list) are found in the Hate class of HS-data. Example, a large HTs-list tends to a low FN. Fact 2: for best precision, the FP is zero. This happens when no HTs (in a HTs-list) are found in the No-Hate class of HS-data. Example, a small HTs-list tends to a low FP. Hence, for a given HS-data, the best conditions to select HTs leads to best precision and recall, thus we can generate a Severe HTs-list.

\textbf{8. Summary\_N(HateTerms):} This provides information of percent HS-lines with N HTs in a HS-data class e.g., x\% have 1 HT, y\% have 2 HTs, z\% have 3 HTs and so on. The results are given in the Table \ref{TabSummary}.

\textbf{Rare instances of the co-occurring HTs} in a hate-speech is also interesting to study. Imbalance occurrences of hate speech as compared to normal speech leads to rare instances of HTs and HS-lines in a HS-data. Such rare HTs and HS-lines are the minority representative of HS-data. We can identify and list those rare HTs by identifying rare concepts and their effect on the classes. It is interesting to analyse those groups of rare HTs (as hate concepts in Section III) and their effect on the classes. It will highlight the concept of a hate speech with a list of those ``rare instances in the HS-data". It is interesting to analyse a rare hate related terms list such as Anglo, yt, obese etc. and their hateful effects on the imbalanced classes.

This section on quantitative analysis leads to the interpretation that it is straight-forward to classify hate content from no-hate content. However, it is hard to classify Hate class from Offensive class because HTs occurs in both contents. Thus, in such conditions, qualitative analysis can provide help in visualization, which is described in next section.
\section{Qualitative analysis: Stable Hate Rules, Concepts, Transitivities, and Lattices}

This section describes qualitative analysis of multiple HS-data. We used ordered sequence of HTs in a HS-data for ordered rule mining, which produces \textit{interesting rules} as compared to the unordered rule mining. Hence, we will use ordered mining, which considers ordering of HTs in a hate speech. This mining is explained and then defined as follows. 

\textbf{Hate Speech Rule Mining Example:} To discover co-occurrences of desired terms, we consider only HTs and contextual terms in a hate speech. Suppose ‘Anglo’ is a contextual term, which is a white English speaking person. Suppose there are 19 tweets (each as a hate speech) with `sp*c', which is an ethnic slur for people from Spanish-speaking. Out of them 3 tweets are as follows
\textbf{Tweet 1:} ``Black cops k*ll white citizens. sp*c cops k*ll Anglo citizens. Z*geuner cops r*pists." \textbf{Tweet 2:} ``No half-breed sp*c Anglo, k*lled so." \textbf{Tweet 3:} ``A*glo-S*xn Protestant, alive US. None, foreign f*lth". The FreqHTs denotes the frequency of a Hate Term (HT) (means number of occurrences of individual HT) in a hate speech. The FreqHT of ‘Anglo’ and ‘sp*c’ are as follows: N(Anglo) = 3 and N(sp*c) = 18. The FreqCoHTs denote the frequency of co-occurring HTs in a hate speech. The FreqCoHTs for (Anglo and sp*c) are as follows: N(Anglo, sp*c) = 2; N(Anglo as antecedent) = 1; and N(sp*c as antecedent) = 15.

When we treat tweets as unordered database, this will result in the following unordered hate rules
\begin{displayquote}
\noindent [Anglo] $\rightarrow$ [sp*c] \#SUP:2 \#CONF: 0.66 means

\noindent N(Anglo $\cup$ sp*c) / N(Anglo) = 2/3 

\noindent [sp*c] $\rightarrow$ [Anglo] \#SUP: 2 \#CONF: 0.11 means

\noindent N(Anglo $\cup$ sp*c) / N(sp*c) = 2/18 
\end{displayquote}

When we treat tweets as ordered sequence database, this will result in the following ordered hate rule

\begin{displayquote}
\noindent [sp*c] $\rightarrow$ [Anglo] \#SUP: 2 \#CONF: 0.13 means

\noindent N(Anglo $\cup$ sp*c) / N(sp*c as antecedent) = 2/15.
\end{displayquote}

\textbf{Definition: Stable Hate Rule (SHR)} mining is performed over multiple Hate Speech data (HS-data) with only hate-terms and Named-entities. This generated Stable Hate Rules (SHRs), which can be read as ``if someone uses a HT ‘A’, then most probably the person may also use HT ‘B’ with a given probability". The SHRs could be like [A] $\rightarrow$ [B], where the [A] is antecedent and the [B] is its consequent.

\begin{figure*}[ht]
\centering
\includegraphics[width=1.0 \linewidth]{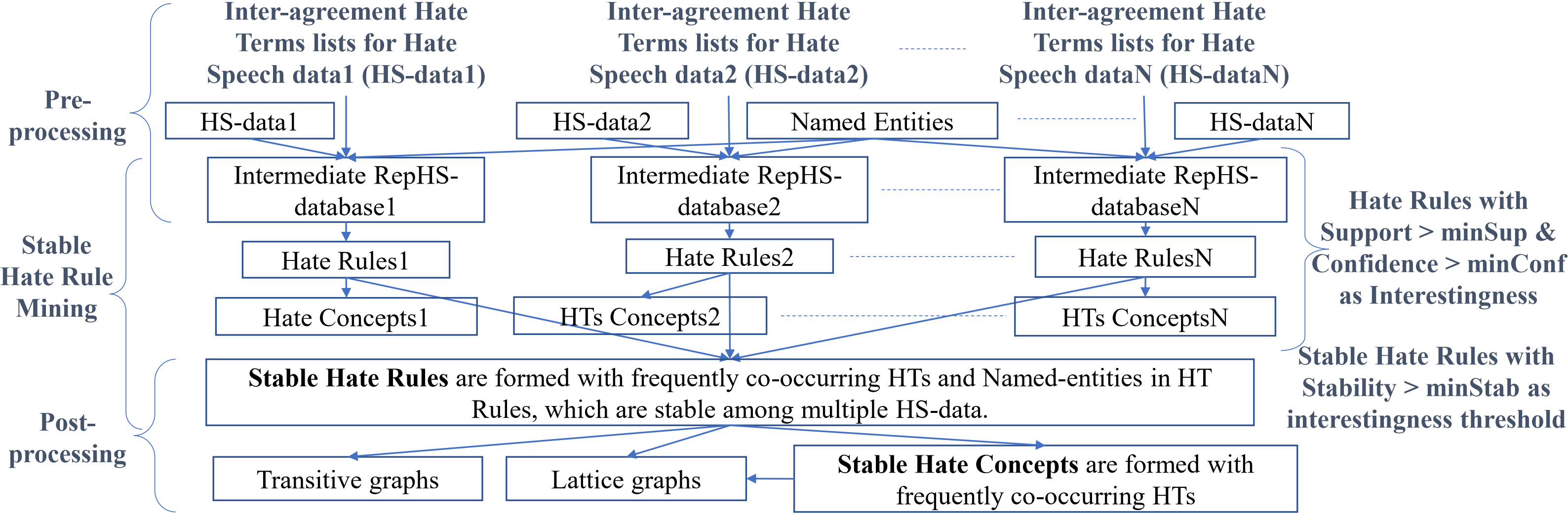}
\vspace{-15pt}
\caption{Methodology for the SHR mining to generate: Stable Hate Rules, Concepts, Transitivities, and Lattices.}
\vspace{-11pt}
\label{Fig3}
\end{figure*}

\textbf{Answer to Q4a:} The Fig. \ref{Fig3} presents an overview of the SHR mining, which has the following four steps.

\textbf{Step 1: Representational Hate Speech Database (RepHS-database)} is an intermediate database created from a given HS-data containing terms of two lists: the Inter-Agreement HT-list and the Named-Entities list. This means the ResHS-database contains each HS line having either hate-terms or named-entities. Each HS-data is pre-processed to RepHS-database containing each row with HTs of the Inter-agreement HTs-list (for corresponding HS-data) and Contextual Keywords (in the Named-Entities list). Semantically, this RepHS-database is a sequential database, which contains a sequence of words containing only HTs (in the Inter-agreement HTs-list) and Named-Entities in each HS-line of the HS-data. This RepHS-database is used for Hate Rule Mining, which applies sequential rule mining (as a special kind of association rule mining) in ordered occurrences of various HTs. 

\textbf{Step 2: Stable Hate Rule (SHR) mining} uses the ordered sequential rule mining over classes (Hate, Relative-hate, No-hate). While SHR mining, the Named Entities is a Context (women, regional, etc) that generates SHRs for some context. The SHR mining over multiple (N) RepHS-database\textbf{s} retrieves several SHRs and concepts. The SHR mining detects the hate rules, which has stability greater than the given threshold of minStab. These hate rules are SHRs, which occur frequently in multiple RepHS-databases. The SHR mining resulted in the following outputs: the ``Hate Rules for each HS-data", the ``Collection of Hate Rules in all HS-data", the ``Outer Join of Hate Rules", and the ``SHRs". For the three classes and multiple HTs-list, each row of the ``Outer Join of Hate Rules" table contains one hate rule with support and confidence. 

\textbf{Step 3: Post-processing visualization} generates hate Concepts, Transitive graphs, and Lattice graphs from similar hate rules or SHRs in the form of $A \rightarrow B$ for two sets of HTs ($A$ and $B$). Following will provide an answer to \textbf{Q4b}:

\textbf{-} A hate Concept is a superset of unique HTs in the similar hate rules that form antecedent and consequent with a set of HTs. Merge HTs in the antecedent and consequent of similar hate rules as a single set of hate Concept without considering the order of rules and the order of HTs in those rules. A hate Concept would be more comprehensive as compared to many similar hate rules. The $A \rightarrow B$ is interpreted as if ordered HTs in A occurs, then it will be followed by the ordered HTs in B. Consider the three HTs $\{a, b, c\}$ that make hate rules: $\{a\}  \rightarrow \{b, c\}$; $\{a\} \rightarrow \{b\}$; $\{a\} \rightarrow \{c\}$. This will produce a hate Concept as $\{abc\}$. 

\textbf{-} We can generate information about the co-occurrences of the HTs, which are represented as SHRs (visualized as Transitive graphs) and Concepts (visualized as Lattice graphs). This aims to detect transitivities and lattices formed between the HTs, which are defined as follows.

\textbf{Definition: Transitive graph} visualizes hate rules with similar Hate-Terms to form a graph such that source nodes are represented by the concatenated HTs in the antecedent and target nodes are represented by the concatenated HTs in the consequent.

\textbf{Definition: Lattice graph} visualizes hate rules with similar Hate-Terms to form a graph such that child nodes are represented by the combination of concatenated HTs of its parent nodes. The root-node is the concatenation of all HTs in all the similar SHRs. 

To construct the graphs, we concatenated HTs in hate rules with the symbol `\_'. Examples of SHRs, concepts, transitivities, and lattices are presented in the Section V.B.
\section{Dataset}

This section describes the 3 Hate Speech data (HS-data) and 6 Hate Terms-lists (HTs-lists). We pre-processed HS-data and the HTs-list before using them for the experimentation (in Section V).

\textbf{1) Hate Speech data (HS-data):} We have used the following three HS-data for the experimentation.

\textit{a) Davidson et al. \cite{davidson2017automated} (Twitter tweets):} The dataset contains 25k tweets with Hatebase lexicon out of 85.4 million tweets. CrowdFlower (CF) workers manually labeled 24,802 out of 25k tweets, according to the given definition of three categories: hate speech, offensive but not hate speech, and neither offensive nor hate speech. Three or more persons done the intercoder-agreement score provided by CF is 92\%. Majority voting is used to label a tweet. Labeling of this dataset is done by random CrowFlower Workers (humans).

\textit{b) de Gibert et al. \cite{de2018hate} (White Supremacy forum):} It is composed of 1000s of sentences extracted from Stormfront, a white supremacist forum. For hate or not hate labelling, an annotation tool was developed that needed manual assistance for choosing the context before labelling: hate or not.

\textit{c) Gao et al. \cite{gao2017detecting} (Fox-news-comments):} Authors created Fox News User Comments Corpus, they annotated corpus of hate speech with context information. The corpus has 1528 annotated comments, out of which 435 labeled as hateful on 10 NEWS comment threads of the Fox News website.

Different HS-data have different class names. Table \ref{Tab2} shows how these dataset classes correspond to our classes (Hate, Relative-hate, No-hate).

\begin{table}[htbp]
\caption{Hate Speech data}\label{Tab2}
\vspace{-8pt}
\begin{center}
\begin{tabular}{|c|c|c|}
\hline
\textbf{Hate Speech}&\multicolumn{2}{|c|}{\textbf{Classes}} \\
\hhline{~--}
\textbf{data} & \textbf{\textit{Used in HS-data}}& \textbf{\textit{Used in our work}} \\
\hline
\hline
& Hate & Hate \\
\hhline{~--}
\multirow{1}{*}{Davidson et al. \cite{davidson2017automated}} & Offensive & Relative-Hate \\
\hhline{~--}
& Non-Offensive & No-Hate \\
\hline
\hline
& Hate & Hate \\
\hhline{~--}
de Gibert et al. \cite{de2018hate} & Relational Hate & Relative-Hate \\
\hhline{~--}
& No-Hate & No-Hate \\
\hline
\hline
& Hate & Hate \\
\hhline{~--}
Gao et al. \cite{gao2017detecting} & -- & Relative-Hate \\
\hhline{~--}
& No-Hate & No-Hate \\
\hline
\end{tabular}
\vspace{-10pt}
\end{center}
\vspace{-7pt}
\end{table}

\textbf{2) Hate Terms-lists (HTs-lists):}
We used the following six sets of HTs-lists, which cover the following hate context.

\textit{a) Chandrasekharan et al.} \cite{chandrasekharan2017you} contains Reddit word list from two subreddits: r/f*tpeoplehate and r/C**nTown, where Reddit hate lexicon\footnote{\scriptsize{https://www.dropbox.com/sh/5ud4fwxvb6q7k20/AAAH\_SN8i5cfmJRKJteEW2b2a}}.

\textit{b) Gorrell et al.} \cite{gorrell2018twits} contains tweets from the UK general elections to explore the abuse directed at politicians. The GATE abuse tagger is available at web-links\footnote{https://cloud.gate.ac.uk/shopfront/displayItem/gate-hate}.

\textit{c) Hatebase}\footnote{https://hatebase.org/academia} uses a broad multilingual vocabulary based on nationality, ethnicity, religion, gender, sexual discrimination, disability and class to monitor incidents of hate speech across 200+ countries. Vocabulary datasets contain valuable lexicon from various data repositories for trend analysis.

\textit{d) Bassignana et al.} \cite{bassignana2018hurtlex} given list named Hurtlex containing multilingual lexicons different targets of hate (immigrants and women) for regional and cultural patterns. HurtLex contains lexicons of hate terms for 50 languages, which are divided into 17 categories, plus a macro-category indicating whether there is stereotype involved)\footnote{https://github.com/valeriobasile/hurtlex}.

\textit{e) Wiegand et al.} \cite{wiegand2018inducing} filtered the abusive words from a given set of negative polar expressions\footnote{https://github.com/uds-lsv/lexicon-of-abusive-words}.

\textit{f) Union:} We made a union list from all the distinct HTs appearing in the above given five HTs-lists available online. 

We found both the HS-data and the given HTs-list has faults in the HTs due to: Stemmer NLP limitations, Not hateful without knowing context, and Annotators agree or disagree (complete consensus). We stemmed the tokenized data both in the HS-data and in the HTs-lists.
\section{Hate Speech Analytics and Experiments}

In this section, first we describe HS-data analysis along with the generation of the Severe HTs-list. Next, we discuss the Hate Rules, Concepts, Transitivities, and Lattice. We also discuss experiments conducted from our Java based implementation consisting of robust and statistical algorithms.

\subsection{Generation of Severe Hate Terms List}
We present the quantitative analysis as we aimed to retrieve Severe HTs-list from the given HTs-lists. The Severe HTs-list produces better results than the given HTs-lists. Although there were many resulting artifacts for each of three HS-data, to keep brevity we present only a few results from Davidson et al. hate class to describe the hate-speech along with the various HTs-lists. We generated 8 artifacts discussed in Section II. The first 5 intra-agreement artifacts are based on Section II.A and next 3 are inter-agreement artifacts based on Section II.B. These 8 artifacts are discussed as follows.

\textbf{1. Creation of hate terms frequencies:} In Table \ref{TabN(X)Terms}, we provided N(0), N(1), ... N(X) HTs found in the HS-lines of the HS-data (Davidson\_et\_al\_ of class 0Hate). The N(0), N(1), ... N(X) hate terms are used from the Severe HTs-list with Offensiveness values greater than threshold \textbf{minimum Offense (minOffense)} = 0.70. This particular example has a maximum value of N = 13 i.e. 13 number of HTs in a single hate speech.

\begin{table}[htbp]
\centering
\vspace{-8pt}
\caption{N(0), N(1), N(2) ... N(X) Terms Example.}\label{TabN(X)Terms}
\vspace{-4pt}
\begin{tabular}{|p{1.1cm}|p{1.2cm}|p{5.1cm}|}
\hline
\textbf{Filename} & \textbf{Hate Term} & \textbf{Tweets}\\

\hline
N(0)\_HTs & --	& \#[IDENTITY] can get a job at the [IDENTITY]. Or as The [IDENTITY]. I hear they like diversity and tolerance. As long as you ain't a cracker \#[TAG] \\

\hline
N(1)\_HTs & f*ggot & @[IDENTITY] answer my [IDENTITY] f*ggot \#[TAG] \\

\hline
N(2)\_HTs & f*ggot; f*ck & @[IDENTITY] f*ck those f*ggots \\
\hline

\hline
so on ... & ...	& ... \\
\hline
\end{tabular}
\vspace{-4pt}
\end{table}

\begin{figure}[hbt!]
\vspace{-7pt}
\centerline{\includegraphics[scale=0.75]{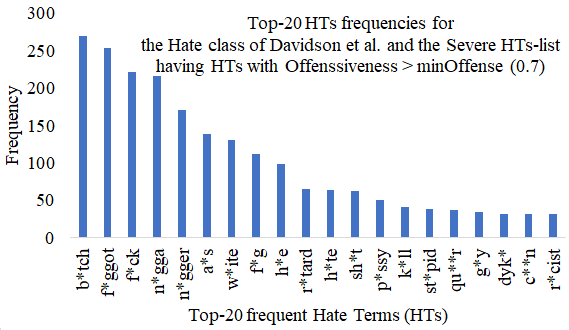}}
\vspace{-7pt}
\caption{Top-20 HTs from Severe HTs-list (0.7) and Davidson et al Hate class.}
\label{Fig4}
\end{figure}

\textbf{2. AllHateTermsFrequencies and TopTermsFrequency:} In Fig. \ref{Fig4}, the count of HTs (belonging to Severe HTs-list) are given for the Hate class of the Davidson et al. (as a Hate class of HS-data) for the top 20 most frequent terms (threshold).

\textbf{3. AllHTsPercentLine:} This signifies X\% of HS-lines with a particular HT in a HS-data class. The artifact tells about the number and percentage of HS-lines in which HT occurs. The Table \ref{TabAllHTsPercentLine} shows first three HTs of Severe HTs-list with Offensiveness $>$ minOffense (0.7) and Hate class of HS-data (Davidson\_et\_al\_0Hate).

\begin{table}[htbp]
\centering
\vspace{-7pt}
\caption{AllHTsPercentLine Example.}\label{TabAllHTsPercentLine}
\vspace{-7pt}
\begin{tabular}{|p{1.3cm}|p{2.4cm}|p{1cm}|p{2.3cm}|}
\hline
\textbf{Hate Term} & \textbf{N(HateTermInLines)} & \textbf{N(Lines)} & \textbf{\%(HateTermLines)} \\
\hline
f*ggot & 249 & 1430 & 17.413 \\
\hline
b*tch & 240 & 1430 & 16.783 \\
\hline
f*ck & 199 & 1430 & 13.916 \\
\hline
so on & ... & ... & ... \\
\hline
\end{tabular}
\end{table}

\textbf{4. OuterJoinHTsFrequencies  and  OuterJoinHTsPercentLines:} The Table \ref{TabOuterJoinHTsFrequencies} and \ref{TabOuterJoinHTsPercentLines} provide examples respectively, for the Severe HTs-list with Offensiveness $>$ minOffense (0.7).

\begin{table}[htbp]
\centering
\vspace{-7pt}
\caption{OuterJoinHTsFrequencies Example.}\label{TabOuterJoinHTsFrequencies}
\vspace{-4pt}
\begin{tabular}{|p{1cm}| p{2cm} | p{2cm} | p{2cm} |}
\hline
\textbf{Hate Term} & \textbf{Davidson et al. 0Hate} & \textbf{Davidson et al 1Offensive} & \textbf{Davidson et al. 2NonOffensive} \\
\hline
f*ggot & 253 & 291 & 1 \\
\hline
b*tch & 269 & 11192 & 11 \\
\hline
f*ck & 221 & 2039 & -- \\
\hline
so on & ... & ... & ... \\
\hline
\end{tabular}
\end{table}

\begin{table}[htbp]
\centering
\vspace{-7pt}
\caption{OuterJoinHTsPercentLines Example.}\label{TabOuterJoinHTsPercentLines}
\vspace{-4pt}
\begin{tabular}{|p{1.2cm}| p{2cm} | p{2cm} | p{2cm} |}
\hline
\textbf{Hate Terms} & \textbf{Davidson et al. 0Hate} & \textbf{Davidson et al 1Offensive} & \textbf{Davidson et al. 2NonOffensive} \\
\hline
f*ggot & 17.413 & 1.501 & 0.024 \\
\hline
b*tch & 16.783 & 54.627 & 0.264 \\
\hline
f*ck & 13.916 & 9.734 & -- \\
\hline
so on & ... & ... & ... \\
\hline
\end{tabular}
\end{table}

\textbf{5. Intra-Agreement-HTs for each HTs-list:} For a particular class of a HS-data, we count the number of HS-lines that contain HTs which belong to a specific HTs-list. For each HT in a HTs-list, we calculated the Hatefulness and Relativeness based on their definitions (in Section II). This information makes the intra-agreement HTs matrix for each given HTs-list. We found a low Relativeness measure when some HT is used alone. In Table \ref{TabIntra-Agreement HTs Matrix} relativeness of `tr*sh' ranges from low as compared to the Relativeness of `eurotr*sh', ``tr**ler park tr*sh'', ``tr**ler tr*sh'', and ``white tr*sh''. 

\begin{table*}
\centering
\caption{Intra-Agreement HTs Example for HS-data (Davidson et al.) and HTs-list (Union).}\label{TabIntra-Agreement HTs Matrix}
\begin{tabular}{| p{1cm} | p{1cm} | p{1.3cm} | p{1cm} | p{1.3cm} | p{1.4cm} | p{1.2cm} | p{1.2cm} | p{1.2cm} | p{1.3cm} | p{1.4cm}|}
\hline
\textbf{Hate Terms (HTs)} & \textbf{Hate Class HS-lines} & \textbf{\#Offensive + Non-Offensive HS-lines} & \textbf{\#Hate Class HS-lines} & \textbf{Hatefulness (Hate Class)} & \textbf{Relativeness (Hate Class)} & \textbf{\#Hate + Offensive HS-lines} & \textbf{Non-Offensive HS-lines} & \textbf{\#Hate + Offensive HS-lines} & \textbf{Hatefulness (Hate + Offensive)} & \textbf{Relativeness (Hate + Offensive)} \\
\hline
f*ggot & 249 & 1 & 1431 & 1 & 0.996 & 537 & 1 & 20622 & 1 & 0.998 \\
\hline
b*tch & 240 & 11 & 1431 & 1 & 0.956 & 10723 & 11 & 20622 & 1 & 0.999 \\
\hline
f*ck & 199 & 0 & 1431 & 1 & 1 & 2067 & 0 & 20622 & 1 & 1 \\
\hline
tr*sh & 106 & 680 & 1431 & 1 & 0.135 & 442 & 680 & 20622 & 1 & 0.394 \\
\hline
eurotr*sh & 0 & 1 & 1431 & 0 & 0 & 1 & 1 & 20622 & 1 & 0.5 \\
\hline
tr**ler park tr*sh & 2 & 1 & 1431 & 1 & 0.667 & 2 & 1 & 20622 & 1 & 0.667 \\
\hline
tr**ler tr*sh & 3 & 2 & 1431 & 1 & 0.6 & 6 & 2 & 20622 & 1 & 0.75 \\
\hline
white tr*sh & 56 & 3 & 1431 & 1 & 0.949 & 91 & 3 & 20622 & 1 & 0.968 \\
\hline
so on ... & ... & ... & ... & ... & ... & ... & ... & ... & ... & ... \\
\hline
\end{tabular}
\end{table*}

\textbf{6. Inter-Agreement-HTs for multiple HTs-lists:} For a HS-data, to make an Inter-agreement HTs among multiple HT-lists, we merged HTs of all the ``Intra-Agreement HTs''. Then, for each HT, we used Hatefulness and Relativeness information to calculate its Offensiveness (as defined in the Section II.B). We found low Offensiveness value of some HTs when used alone or without any context e.g. offensiveness of 'tr*sh' is low as compared to the offensiveness of `eurotr*sh', ``tr**ler park tr*sh'', ``tr**ler tr*sh'', and ``white tr*sh''. The Table \ref{TabInter-Agreement HTs} shows the three HTs for the inter-agreement between the Davidson et al. tweets and the given six HT-lists.

\textbf{Answer to RQ2:} Table \ref{TabInter-Agreement HTs} shows high Offensiveness results in optimum number of HTs. For each HT, Hatefulness and Relativeness in Table \ref{TabInter-Agreement HTs} is taken from the Hatefulness and Relativeness (shown in Table \ref{TabIntra-Agreement HTs Matrix}). Offensiveness of Table \ref{TabInter-Agreement HTs} is calculated with the Hatefulness and Relativeness. Using Offensiveness value, we discriminate severe and mild hate lexicons, which created a Severe HTs-list to classify hate speech. This means, for high Offensiveness, the selected highly severe HTs can help to perform better hate speech classification.

\begin{table*}
\centering
\caption{Inter-Agreement HTs between the Davidson et al. and the six given HTs-lists.}\label{TabInter-Agreement HTs}
\begin{tabular}{|p{1.5cm}|p{1.3cm}|p{1.3cm}|p{1.5cm}|p{2cm}|p{2cm}|p{2cm}|p{3cm}|} 
\hline
\textbf{HTs} &	\textbf{Hatefulness (Hate)} & \textbf{Relativeness (Hate)} & \textbf{Offensiveness (Hate)} & \textbf{Hatefulness (Hate+Offensive)} & \textbf{Relativeness (Hate+Offensive)} & \textbf{Offensiveness (Hate+Offensive)} & \textbf{HateListNames} \\
\hline
f*ggot & 1 & 0.996 & 0.998 & 1 & 0.998 & 0.999 & Chandrasekharan et al Reddit hate lexicon; Gorrell et al abuse-terms; HateBaseList; hurtlex\_EN; Union; Wiegand et al \\
\hline
b*tch & 1 & 0.956 & 0.978 & 1 & 0.999 & 0.999 & Gorrell et al abuse-terms; HateBaseList; hurtlex\_EN; Union; \\
\hline
f*ck & 1 & 1 & 1 & 1 & 1 & 1 & hurtlex\_EN; Union; Wiegand et al \\ 
\hline
tr*sh & 1 & 0.135 & 0.238 & 1 & 0.394 & 0.565 & HateBaseList; hurtlex\_EN; Union \\
\hline
eurotr*sh & 0 & 0 & NaN & 1 & 0.5 & 0.667 & HateBaseList; Union \\
\hline
tr**ler park tr*sh & 1 & 0.667 & 0.8 & 1 & 0.667 & 0.8 & HateBaseList; Union \\
\hline
tr**ler tr*sh & 1 & 0.6 & 0.75 & 1 & 0.75 & 0.857 & HateBaseList; Union \\
\hline 
white tr*sh & 1 & 0.949 & 0.974 & 1 & 0.968 & 0.984 & HateBaseList; Union \\
\hline
so on ... & ... & ... & ... & ... & ... & ... & ... \\
\hline
\end{tabular}
\end{table*}

\textbf{7. Inter-agreement Confusion-matrix:} We intended to generate a Severe HTs-list as presented in Section II.B. This provides comparison between the Severe HTs-list and the baseline HTs-lists (Chandrasekharan et al., Gorrell et al., Hatebase, Hurtlex, and Wiegand et al.). All these HTs-lists have inter-agreement between the HS-data and HTs. First, as given in the Table \ref{Tab1}, we constructed the confusion-matrix (with TP, TN, FP, FN) to calculate the precision and recall. Then, we calculate the accuracy, precision, recall, and f-measure (or f-score). 

\textbf{Answer to RQ3:} The Table \ref{TabConfusionMatrix} shows the comparison between the Severe HTs-list and a given HTs-list which is best among all HTs-lists. In all cases, the Severe HTs-list resulted in best F-Score and contains best inter-agreeing HTs for a HS-data. This is because the Severe HTs-list contains the best HTs of all the HTs-list.

Table \ref{TabConfusionMatrix} shows results for high offensiveness threshold minOffense to make Severe HTs-list, which performed better than the classification provided by all the given set of HTs-lists. We found that a Severe HTs-list distinguishes between hateful and non hateful lines to achieve better classification.

Table \ref{TabConfusionMatrix} presents comparisons for the three HS-data. First, to classify Davidson et al. Hate Vs No-hate class, we generated a Severe HTs-list with 298 HTs with a Offensiveness greater than minOffense (0.7). The Severe HTs-list resulted in the F-measure of 0.923, which is better as compared to the HS-list of Gorrell et. al. which contains 403 abuse-terms and resulted in the 0.845 F-measure (best among all given HTs-lists). Second, for HS-data of DeGilbert et al, we generated a Severe HTs-list with 578 HTs with a Offensiveness greater than minOffense (0.46). The Severe HTs-list perform better as compared to the HS-list Union (13538) that produced best F-measure among the given HTs-lists. Third, for HS-data of Goa et al, we generated a Severe HTs-list with 622 HTs with a Offensiveness greater than minOffense (0.75). The Severe HTs-list performs better as compared to HS-list Union (13538) that produced better results among the given HTs-lists.

\begin{table*}
\caption{For the three HS-data, the table provides a comparison of the Severe HTs-list with the given HTs-lists. \textbf{This answered RQ3}}\label{TabConfusionMatrix}
\centering
\begin{tabular}{| p{2.8cm} | p{6.3cm} | c | c | c | c | p{1.6cm} |}
\hline
\textbf{HTs-list Name (minOffense, number of HTs)} & \textbf{HS-data Name and Class} & \textbf{Accuracy} & \textbf{Recall} & \textbf{Precision} & \textbf{F-Measure} & \textbf{Compute Time} \\
\hline
\hline
Gorrell et al abuse-terms & Davidson\_et\_al\_ 0Hate Vs. No-Hate & 0.857 & 0.784 & 0.917 & 0.845 & \\
\hhline{~-----}
(-, 403) & Davidson\_et\_al\_ 0Hate+1Offensive Vs. No-Hate & 0.845 & 0.761 & 0.915 & 0.831 & 12 sec\\
\hhline{~-----}
& Davidson\_et\_al\_1 Offensive Vs. No-Hate & 0.844 & 0.759 & 0.915 & 0.83 & \\
\hline
\textbf{Offensiveness(Hate)} & Davidson\_et\_al\_ 0Hate Vs. No-Hate & \textbf{0.921} & \textbf{0.946} & 0.901 & \textbf{0.923} & \\
\hhline{~-----}
\textbf{(0.7, 298)} & Davidson\_et\_al\_ 0Hate+ 1Offensive Vs. No-Hate & \textbf{0.929} & \textbf{0.962} & 0.903 & \textbf{0.931} & 17 sec \\
\hhline{~-----}
& Davidson\_et\_al\_1 Offensive Vs. No-Hate & \textbf{0.93} & \textbf{0.963} & \textbf{0.903} & \textbf{0.932} & \\
\hline
\hline
& de\_Gibert\_et\_al\_ 0Hate Vs. No-Hate & 0.633 & 0.959 & 0.58 & 0.723 & \\
\hhline{~-----}
Union (-, 13538) & de\_Gibert\_et\_al\_0Hate +1RelationalHate Vs. No-Hate & 0.629 & 0.951 & 0.578 & 0.719 & 1 min 31 sec\\
\hhline{~-----}
& de\_Gibert\_et\_al\_ 1RelationalHate Vs. No-Hate & 0.6 & 0.893 & 0.563 & 0.69 & \\
\hline 
\textbf{Offensiveness(Hate)} & de\_Gibert\_et\_al\_ 0Hate Vs. No-Hate & \textbf{0.821} & 0.832 & \textbf{0.814} & \textbf{0.823} & \\
\hhline{~-----}
\textbf{(0.46, 578)} & de\_Gibert\_et\_al\_ 0Hate+ 1RelationalHate Vs. No-Hate & \textbf{0.8} & 0.789 & \textbf{0.806} & \textbf{0.797} & \textbf{14 sec}\\
\hhline{~-----}
& de\_Gibert\_et\_al\_ 1RelationalHate Vs. No-Hate & \textbf{0.646} & 0.482 & \textbf{0.718} & 0.577 & \\
\hline
\hline
Union (-, 13538) & Gao\_et\_al\_ 0Hate Vs. No-Hate & 0.46 & 0.772 & 0.475 & 0.588 & 15 sec\\
\hline 
\textbf{Offensiveness(Hate) (0.75, 622)} & Gao\_et\_al\_ 0Hate Vs. No-Hate & \textbf{0.541} & 0.718 & \textbf{0.53} & \textbf{0.61} & \textbf{5 sec} \\
\hline
\end{tabular}
\end{table*}

\begin{table}
\vspace{-4pt}
\caption{Ranking of HTs-list Name in decreasing order of Inter-agreement with the HS-data.}\label{TabHTs-listRanking}
\centering
\begin{tabular}{| p{1.4cm} | p{6.6cm}|}
\hline
\textbf{HS-data Name} & \textbf{HTs-lists Names (number of HTs)} \\
\hline
\hline
\textbf{Davidson et al} & \textbf{Offensiveness(Hate) (0.7, 298)}, Gorrell et al abuse-terms (403), HateBaseList (1015), Wiegand et al lexicon-of-abusive-words (7156), Hurtlex EN (5925), Union (13538), and Chandrasekharan et al Reddit hate lexicon (199). \\
\hline
\textbf{de Gibert et al} & \textbf{Offensiveness(Hate)(0.46, 578)}, Union (13538), Hurtlex EN (5925), Wiegand et al lexicon-of-abusive-words (7156), Chandrasekharan et al Reddit hate lexicon (199), HateBaseList (1015), Gorrell et al abuse-terms (403). \\
\hline
\textbf{Gao et al} & \textbf{Offensiveness(Hate)(0.75, 622)}, Union(13538), Wiegand et al lexicon-of-abusive-words (7156), Hurtlex EN (5925), Chandrasekharan et al Reddit hate lexicon (199), Gorrell et al abuse-terms (403), HateBaseList(1015). \\
\hline
\end{tabular}
\vspace{-10pt}
\end{table}

To produce a high value of precision and recall, we used two steps. First, we added HTs having high-severity (i.e., HTs occurring mostly in the Hate class), which have highest offensiveness. Second, we ignored HTs having less-severity (i.e., HTs appearing frequently in the Non-Offensive class), which has low offensiveness. The optimized size of Severe HTs-list contains only best HTs with high offensiveness, which produce best classification.

For HS-data of Davidson et al, we used six given HTs-lists to retrieve a Severe HTs-lists. Instead of using all terms, we retrieved a subset of highly severe terms by setting an minOffense = 0.7, which produced an optimized number of 298 HTs. These HTs are used to classify HS-line in a HS-data as hateful if they contain one or more hate terms. Similarly, we retrieved Severe HTs-list for the other HS-data: de\_Gilbert et al. and Gao et al.

Table \ref{TabHTs-listRanking} shows the ranking of the HTs-list, which provides comparison of inter-agreement between a HS-data with HTs-lists (with number of HTs). The ranking is ordered based on inter-agreement from high to low. From the table, we inferred that ``inter-agreement does not depend on the number of HTs'', but ``inter-agreement depends upon the agreement of HTs in HTs-list with the classes of HS-data''. We also inferred that ``a HTs-list may well inter-agreed with a HS-data and may not be well inter-agreed with another HS-data''.

\begin{table*}[ht]
\caption{For the three HS-data and six HTs-list, the table provide summarised overview.}\label{TabSummary}
\centering
\begin{tabular}{| c | c | c | c | c | c |}
\hline
\textbf{Dataset Name and Class}	& \textbf{HateList Name} & \textbf{HateTerms(N)} & \textbf{N(Entries)}	& \textbf{TotalLines} & \textbf{\%(Entries)} \\
\hline
Davidson et al 0Hate & Chandrasekharan et al Reddit hate lexicon & 0 & 581 & 1430 & 40.629 \\
\hline
Davidson et al 0Hate & Chandrasekharan et al Reddit hate lexicon & 1 & 671 & 1430 & 46.923 \\
\hline
so on ... & ... & ... & ... & ... & ... \\
\hline
Davidson et al 1Offensive & Chandrasekharan et al Reddit hate lexicon & 0 & 16101 & 19190 & 83.903 \\
\hline
Davidson et al 1Offensive & Chandrasekharan et al Reddit hate lexicon & 1 & 2654 & 19190 & 13.83 \\
\hline
so on ... & ... & ... & ... & ... & ... \\
\hline
\end{tabular}
\vspace{-9pt}
\end{table*}

\textbf{8. Summary\_N(HateTerms):} Table \ref{TabSummary} summaries following information about the HS-data and HTs-list. The number of HTs ``HateTerms(N)'' in HS-lines is depicted by N(Entries). Table shows a few rows for the three classes \{0Hate, 1Offensive, 2NonOffensive\} of Davidson\_et\_al and the HTs-list Chandrasekharan\_et\_al. Total number of rows has the N(Entries) $\times$ the number of classes in a HS-data $\times$ the number of given HTs-lists. Similar tables can be made for HS-data: de\_Gibert\_et\_al\_ and Gao\_et\_al.

\subsection{Stable Hate Rules, Concepts, Transitivities, and Lattices}
\label{ssec:layout}
Based on Section III, we present the qualitative analysis to visualize the Stable Hate Rules (SHRs) that can provide information about the co-occurrences of the rare HTs in HS-data. The SHRs are represented and visualized as Transitive graphs. In addition, the hate Concepts are being created from Stable Hate Rules (SHRs). These hate Concepts are visualized as Lattice graphs. \textbf{(This subsection answers to Q4b.)}

Basically, the rule-mining on HS-data suffers with two limitations. First, rule-mining over full HS-data generated exhaustive number of rules having hate-terms along with the stop-words (e.g. In The To a, Can I, You n*gga, As a, etc.), which makes less sense. Thus, we removed stop-words from HS-data. Second, from the HS-data without stop-words, we made three different classes (hate, relative-hate, no-hate) that again resulted in an exhaustive number of rules. Finally, we made an intermediate Representational HS Database (RepHS-database) from Hate and Relative-hate class only. We made a RepHS-database with sequences of each hate speech containing only HTs and contextual Named-Entities, which resulted in patterns of co-occurring HTs in a context.

We made total 9 RepHS-database for mining from \{Davidson et al. (hate and offensive), de Gilbert et al. (hate and relational-hate), Gao et al.(hate)\} $\times$ their three Inter-agreement HTs-lists. In each RepHS-Database, we kept hate speech with only HTs in the three Inter-agreement HTs-lists and Named Entity (list of synonyms for Women (like girl, wife etc.) and Regions (like name of states or countries\footnote{https://meta.wikimedia.org/wiki/List\_of\_countries\_by\_regional\_classification}). This provides rules related to hate speech against Women and Regions.

For the better interpretation from rules, we also included a number of occurrences of SHR's antecedent, support, and confidence. We used association rule mining and sequential rule mining outputs with multiple threshold values of minimum support (minSup) and minimum confidence (minConf). For conciseness, we omitted such detail about support and confidence. For sequential rule mining, we used the RuleGrowth algorithm \cite{fournier2015mining} given in the SPMF data mining tools. The SHR mining is an extension of earlier work \cite{{chaturvedi2019minstab}} and it is useful for the Hate Speech Project.

\begin{figure}[ht]
\centering
\includegraphics[width=0.45 \linewidth]{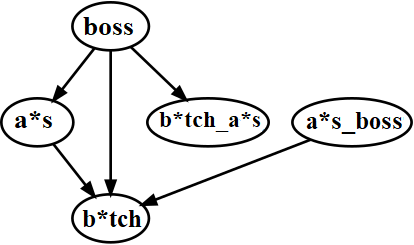}
\includegraphics[width=0.53 \linewidth]{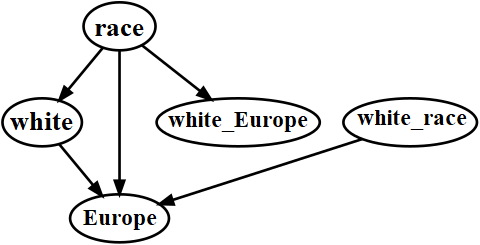}
\caption{Two Transitivity graphs of two sets of similar SHRs: ``a*s\_b*tch\_boss'' and ``Europe\_race\_white'' for Women and Regional context respectively.}
\label{Fig5}
\end{figure}

\begin{figure}[ht]
\centering
\includegraphics[width=0.45 \linewidth]{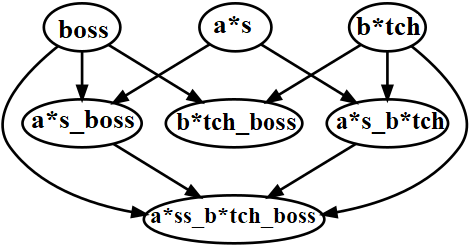}
\includegraphics[width=0.53 \linewidth]{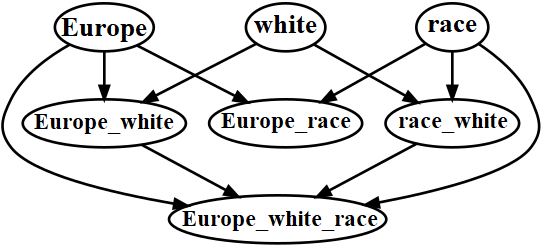}
\caption{Two Lattice graphs of two Concepts: ``a*s\_b*tch\_boss'' and ``Lattice\_Europe\_race\_white''. Root-nodes (at the bottom) are the hate Concepts.}
\label{Fig6}
\end{figure}

Table \ref{Tab5} demonstrates two hate concepts generated from similar SHRs with the same set of HTs. The rules resulted in the formation of two transitive graphs as shown in Fig. \ref{Fig5}. The rules also resulted in the formation of two lattice graphs of two concepts as shown in Fig. \ref{Fig6}. There are many such transitive and lattice graphs, which we skip presenting to keep brevity.

\begin{table}[htbp]
\caption{Two hate concepts (first row) and their SHRs with similar HTs.} \label{Tab5}
\centering
\begin{tabular}{|p{4cm}|p{4cm}|}
\hline
\textbf{a*s b*tch boss 5} 

a*s $\rightarrow$ b*tch

boss $\rightarrow$ b*tch a*s

a*s boss $\rightarrow$ b*tch

boss $\rightarrow$ b*tch

boss $\rightarrow$ a*s
&

\textbf{Europe race white 5}

white $\rightarrow$ Europe

race $\rightarrow$ white Europe

white race $\rightarrow$ Europe

race $\rightarrow$ white

race $\rightarrow$ Europe \\
\hline
\end{tabular}
\vspace{-7pt}
\end{table}

\vspace{-4pt}
\vspace{-7pt}

\vspace{-2pt}
\vspace{-5pt}
\section{Related Works}

Caselli et al. \cite{caselli2020feel} described abuse and offense by studying Inter-Agreement. We have also analyzed Inter-agreement between HTs-list and HS-data Classes, such that, this process generates Severe HTs-list. Pedersen \cite{pedersen2019duluth} presented lists of HTs, whereas we have retrieved Severe HTs-list.

Liu et al. \cite{liu2019improving} presented Multi-Task Deep Neural Network (MT-DNN) for learning text representations for multiple natural language understanding (NLU). Zampieri et al. \cite{zampieri2019predicting}\cite{zampieri2019semeval} annotated the datasets for abusive messages named Offensive Language Identification Dataset (OLID), which they further used for Offensive Language in Social Media (OffensEval). Devlin et al. \cite{devlin2018bert} presented a popular language representation model named Bidirectional Encoder Representations from Transformers (BERT) for tasks like question answering and language inference.

Recently, Mittos et al. \cite{mittos2020and} measured hate and toxicity with the percentage of hate words, and the level of toxicity/inflammatory. They used HTs available on hatebase.org, which were previously worked-out by Hine et al. \cite{hine2017kek}.  Vidgen et al. \cite{vidgen2020learning} presented dynamic human and model based four rounds of hate speech training. Ball et al. \cite{ball2021differential} studied racial dialect using neural networks for harmful tweet detection. Vidgen et al. \cite{vidgen2021introducing} described a contextual abuse dataset with six taxonomies for abusive, non-abusive, and neutral speech.
\section{Conclusions}
To collect inter-agreement information about the HTs-list (Hate Terms list) and the HS-data (Hate Speech data), we answered the four research questions. We generated interesting reports that include: top frequent HTs, intra/inter-agreement of HTs in HTs-list with the HS-data, summarized hate-term occurrences, and Offensiveness of HTs. We also retrieved Stable Hate Rules (SHRs) and Concepts that are used to identify lattices and transitivity relationships between various HTs for a context. For quantitative analysis, using the proposed threshold minOffense for HTs, our Severe HTs-list has out-performed all the given HTs-lists. For qualitative analysis, our SHRs provided visual analytic as Transitive and Lattice graphs of the HTs co-occurring in HS-data for context of Women and Regions.

\section*{Acknowledgments}
Thanks to \textit{Prof. Nishanth Sastry}, \textit{Dr. Bertie Vidgen}, and \textit{Dr. Jatinder Singh} for their discussions on the topic. Authors also thank \textit{The Alan Turing Institute} for supporting this project by providing fellowship to Dr. Animesh Chaturvedi as Post Doctoral - RA at \textit{King’s College London} (U.K.).

\end{document}